\documentclass[10pt,journal,cspaper,compsoc]{IEEEtran}
\usepackage{graphicx}
\usepackage{subfigure}
\usepackage{array}
\usepackage{amsmath,epsfig}
\usepackage{booktabs}
\usepackage[dvipdfm,
CJKbookmarks=true, bookmarksnumbered=true, bookmarksopen=true,
colorlinks=true,
pdfborder=001,
citecolor=blue, linkcolor=blue, anchorcolor=red, urlcolor=black
]{hyperref}
\begin{document}\sloppy

\title{Graphical Representation for Heterogeneous Face Recognition}

\author{Chunlei~Peng,~Xinbo~Gao,~\IEEEmembership{Senior Member, IEEE},~Nannan~Wang,~\IEEEmembership{Member, IEEE},~and~Jie~Li~
\IEEEcompsocitemizethanks{\IEEEcompsocthanksitem C. Peng and J. Li are with the Video and Image Processing System Laboratory, School of Electronic Engineering, Xidian University, Xi'an 710071, Shaanxi, P. R. China.\protect\\ E-mail: clp.xidian@gmail.com; leejie@mail.xidian.edu.cn
\IEEEcompsocthanksitem X. Gao is with the State Key Laboratory of Integrated Services Networks, School of Electronic Engineering, Xidian University, Xi'an 710071, Shaanxi, P. R. China.\protect\\
E-mail: xbgao@mail.xidian.edu.cn
\IEEEcompsocthanksitem N. Wang is with the State Key Laboratory of Integrated Services Networks, School of Telecommunications Engineering, Xidian University, Xi'an 710071, Shaanxi, P. R. China.\protect\\
E-mail: nnwang@xidian.edu.cn.}}

\IEEEcompsoctitleabstractindextext{%
\begin{abstract}

Heterogeneous face recognition (HFR) refers to matching face images acquired from different sources (\textit{i.e.}, different sensors {or} different wavelengths) for identification. HFR plays an important role in both biometrics research and industry. In spite of promising progresses achieved in recent years, HFR is still a challenging problem due to the difficulty to represent two heterogeneous images in a homogeneous manner. Existing HFR methods either represent an image ignoring the spatial information, or rely on a transformation procedure which complicates the recognition task. Considering these problems, we propose a novel graphical representation based HFR method (G-HFR) in this paper. Markov networks are {employed} to represent heterogeneous image patches separately, which {takes} the spatial compatibility between neighboring image patches into consideration. A coupled representation similarity metric (CRSM) is designed to measure the similarity between obtained graphical representations. Extensive experiments conducted on multiple HFR scenarios (viewed sketch, forensic sketch, near infrared image, and thermal infrared image) show that the proposed method outperforms state-of-the-art methods.

\end{abstract}

}

\maketitle
\IEEEdisplaynotcompsoctitleabstractindextext
\IEEEpeerreviewmaketitle

\section{Introduction}
\label{section I}

\IEEEPARstart{F}{}ace images captured through different sources, such as sketch artists and infrared imaging devices, are called in different modalities, \textit{i.e.} heterogeneous {face images}. Matching face images in different modalities, which is referred as heterogeneous face recognition (HFR), is now attracting growing attentions in both biometrics research and industry. For instance, there are circumstances where the photo of the suspect is not available and matching sketches to a large-scale database of mug shots is desired; Matching near infrared (NIR) images or thermal infrared (TIR) images to visual (VIS) images is important for biometric security control to handle complicated illumination conditions.

Because of the great discrepancies between heterogeneous face images, conventional homogeneous face recognition methods perform poorly by directly identifying the probe image (\textit{e.g.} face sketch or infrared image) from gallery images (\textit{e.g.} face photos). Existing approaches can be generally grouped into three categories: {synthesis-based} methods, common space projection based methods, and feature descriptor based methods. {Synthesis-based} methods \cite{Ref3,Ref2,Ref1,Ref6,Ref4,Ref46} first transform the heterogeneous face images into the same modality (\textit{e.g.} photo). Once the synthesized photos are generated from non-photograph images or vice versa, conventional face recognition algorithms can be applied directly. However, the synthesis process is actually more difficult than recognition and the performance of these methods heavily depends on the fidelity of the synthesized images. Common space projection based methods \cite{Ref11,Ref13,Ref8,Ref10,Ref9,Ref12} attempt to project face images in different modalities into a common subspace where the discrepancy is minimized. Then heterogeneous face images can be matched directly in this common subspace. Yet the projection procedure always causes information loss which decreases the recognition performance. Feature descriptor based methods \cite{Ref19,Ref20,Ref22,Ref15,Ref18} first represent face images with local feature descriptors. These encoded descriptors can then be utilized for recognition. However, most existing methods of this category represent an image ignoring the special spatial structure of faces, which is crucial for face recognition in reality.

This paper proposes a novel graphical representation based HFR approach (G-HFR), which does not rely on any synthesis or projection procedure but takes spatial information into consideration. After face images are divided into overlapping patches, Markov networks are {employed} to model the relationship between homogeneous image patches based on a representation dataset. The representation dataset consists of a number of heterogeneous face image pairs. Then the weight matrixes generated from the Markov networks are regarded as graphical representations, which are irrelevant to heterogeneity. Therefore, the similarity between the weight matrixes of heterogeneous face images is used for matching. Considering the spatial structure between heterogeneous face image patches, a coupled representation similarity metric (CRSM) is designed to measure the similarity between their graphical representations. Finally, calculated similarity scores between heterogeneous face images are applied for recognition.

The performance of the proposed G-HFR approach is thoroughly validated on four HFR scenarios: the viewed sketch database (the CUHK Face Sketch FERET Database (CUFSF) \cite{Ref18}), the forensic sketch database (IIIT-D Sketch Database \cite{Ref41}, PRIP Viewed Software-Generated Composite Database (PRIP-VSGC) \cite{Ref42}, our collected forensic sketch database), the near infrared database (the CASIA NIR-VIS 2.0 Face Database \cite{Ref43}), and the thermal infrared database (the Natural Visible and Infrared facial Expression Database (USTC-NVIE) \cite{Ref44}). Experimental results illustrate that the proposed approach achieves superior performance in comparison to state-of-the-art methods.

The main contributions of this paper are summarized as follows:
\begin{enumerate}
    \item We employ Markov networks to obtain graphical representations for representing heterogeneous face images, which firstly takes spatial information into consideration;
    \item A coupled representation similarity metric is developed for matching, which considers the spatial structure between heterogeneous face image patches;
    \item Leading accuracies are achieved on multiple HFR scenarios which illustrates the effectiveness of the proposed method.
\end{enumerate}

{In this paper, excepted when noted, a bold lowercase letter denotes a column vector and a bold uppercase letter {stands for} a matrix. The regular lowercase and uppercase letters denote scalars.} The organization of the rest of this paper is as follows. Section \ref{section II} gives a review on representative HFR methods. Section \ref{section III} presents the proposed graphical representation approach for HFR. Section \ref{section IV} shows the experimental results and analysis and the conclusion is drawn in Section \ref{section V}.

\section{Related Work}
\label{section II}

In this section, we briefly review representative HFR methods in aforementioned three categories: {synthesis-based} methods, common space projection based methods, and feature descriptor based methods.

{Synthesis-based} HFR methods began with an eigen-transformation algorithm \cite{Ref1} proposed by Tang and Wang. Later, Liu \textit{et al.} \cite{Ref2} proposed a locally linear embedding approach for patch-based face sketch synthesis. The sketch patches were synthesized independently and the spatial compatibility between neighboring patches was neglected. Chen \textit{et al.} \cite{Ref7} proposed to learn the locally linear mappings between NIR and VIS patches in a similar manner as \cite{Ref2}. Gao \textit{et al.} \cite{Ref3} employed embedded hidden Markov model to represent the non-linear relationship between sketches and photos and a selective ensemble strategy \cite{Ref39} was explored to synthesize a sketch. Wang and Tang \cite{Ref4} proposed a multi-scale Markov random field model for face sketch-photo synthesis, which {takes} the spatial constraints between neighboring patches into consideration. Li \textit{et al.} \cite{Ref46} proposed a {learning-based} framework to synthesize photos from thermal infrared images and the Markov random field model was applied to improve the synthesized result. Zhou \textit{et al.} \cite{Ref5} proposed a Markov weight field model which was capable of synthesizing new patches that do not appear in the training set. Wang \textit{et al.} \cite{Ref6} presented a transductive face sketch-photo synthesis method which {incorporates} the test image into the learning process.

In order to minimize the intra-modality difference, Lin and Tang \cite{Ref8} proposed a common discriminant feature extraction (CDFE) approach to map heterogeneous features into a common feature space. The canonical correlation analysis (CCA) was applied to learn the correlation between NIR and VIS face images by Yi \textit{et al.} \cite{Ref12}. Lei and Li \cite{Ref13} proposed a subspace learning framework for heterogeneous face matching, which {is} called coupled spectral regression (CSR). They later improved the CSR by learning the projections based on all samples from all modalities \cite{Ref14}. Sharma and Jacobs \cite{Ref9} used partial least squares (PLS) to linearly map images from different modalities to a common linear subspace. A cross modal metric learning (CMML) algorithm was proposed by Mignon and Jurie \cite{Ref10} to learn a discriminative latent space.  Both the positive and negative constraints were considered in metric learning procedure. Kan \textit{et al.} \cite{Ref11} proposed a multi-view discriminant analysis (MvDA) method to obtain a discriminant common space for recognition. The correlations from both inter-view and intra-view were exploited.

A number of feature descriptor based HFR approaches have shown promising performances. Klare \textit{et al.} \cite{Ref15} proposed a local feature-based discriminant analysis (LFDA) framework through scale invariant feature transform (SIFT) feature \cite{Ref16} and multiscale local binary pattern (MLBP) feature \cite{Ref17}. A face descriptor based on coupled information-theoretic encoding was designed for matching face sketches with photos by Zhang \textit{et al.} \cite{Ref18}. The coupled information-theoretic projection tree was introduced and was further extended to the randomized forest with different sampling patterns. Another face descriptor called local radon binary pattern (LRBP) was proposed in \cite{Ref19}. The face images were projected onto the radon space and encoded by local binary patterns (LBP). A histogram of averaged oriented gradients (HAOG) face descriptor was proposed to reduce the modality difference \cite{Ref20}. Lei \textit{et al.} \cite{Ref25} proposed a discriminant image filter learning method benefitted from LBP like face representation for matching NIR to VIS face images. Alex \textit{et al.} \cite{Ref23} proposed a local difference of Gaussian binary pattern (LDoGBP) for face recognition across modalities.

With great progresses achieved on viewed sketches, recently researches began to focus on matching forensic sketches to mug shots. Klare \textit{et al.} \cite{Ref15} matched forensic sketches to mug shot photos with a populated gallery. Bhatt \textit{et al.} \cite{Ref21} proposed a discriminative approach for matching forensic sketches to mug shots {employing} multi-scale circular Weber's local descriptor (MCWLD) and an evolutionary memetic optimization algorithm. Klare and Jain \cite{Ref22} represented heterogeneous face images through their nonlinear kernel similarities to a collection of prototype face images. Considering the fact that many law enforcement agencies employ facial composite software to create composite sketches, Han \textit{et al.} \cite{Ref24} proposed a component based approach for matching composite sketches to mug shot photos.

\section{Graphical Representation for Heterogeneous Face Recognition}
\label{section III}

\begin{figure*}
\begin{center}
    \includegraphics[width=0.8\linewidth]{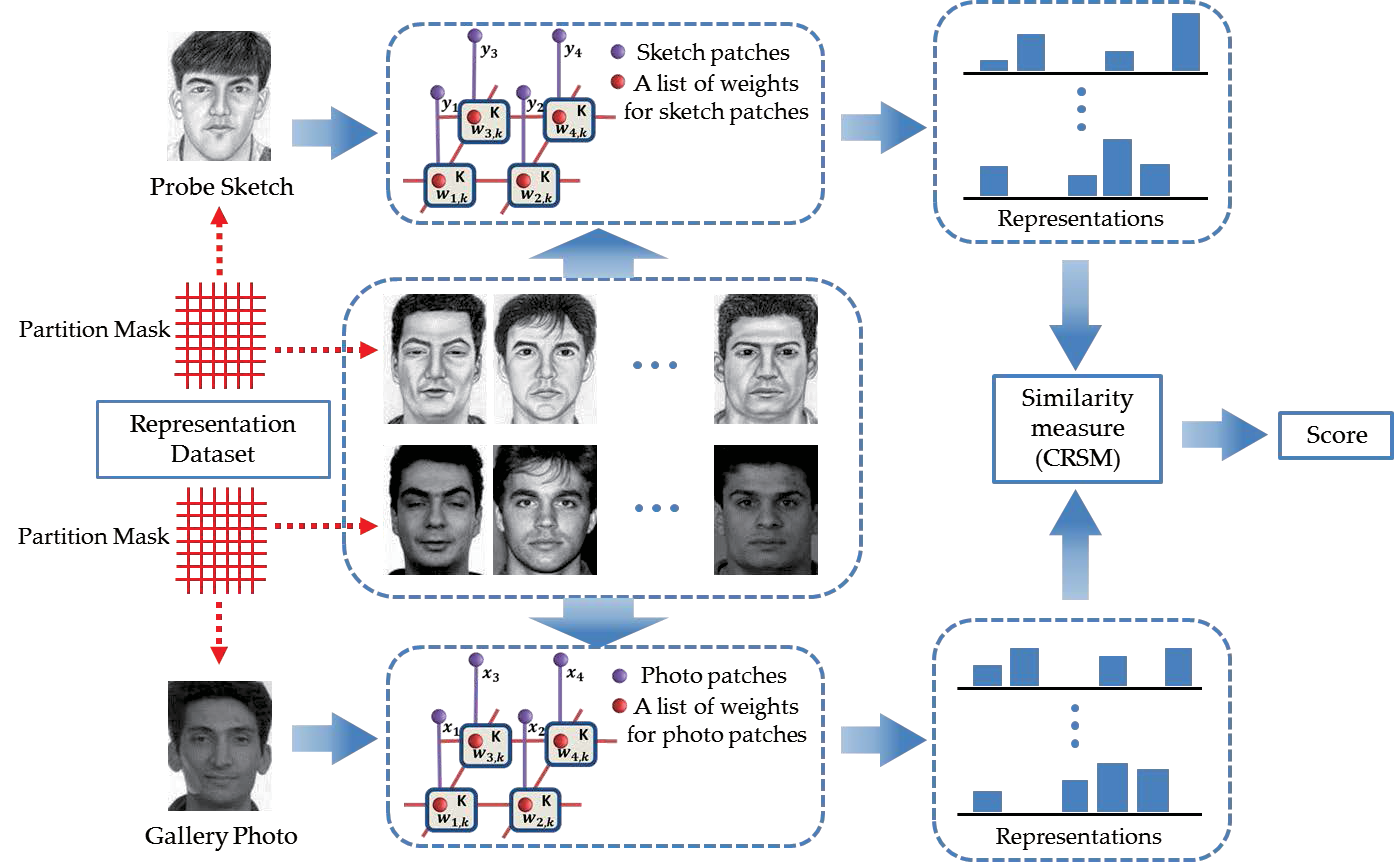}
\end{center}
   \caption{Overview of the proposed graphical representation based heterogeneous face recognition.}
\label{Figure1}
\end{figure*}

In this section, we present a new approach for HFR. Without loss of generality and for ease of representation, we take face sketch-photo recognition as an example to describe the proposed method. {A representation dataset composed of face sketch-photo pairs is constructed in the begining, which is utilized to extract the graphical representations of the gallery and probe images.} Considering a representation dataset with $M$ face sketch-photo pairs $\{(\mathbf{s}^1,\mathbf{p}^1),\cdots,(\mathbf{s}^M,\mathbf{p}^M)\}$, we first divide each face image into $N$ overlapping patches. The probe sketch $\mathbf{t}$ and the gallery photos $\{\mathbf{g}^1,\cdots,\mathbf{g}^L\}$ are also divided into $N$ overlapping patches {correspondingly}. Here $L$ denotes the number of photos in the gallery. For a probe sketch patch $\mathbf{y}_i$($i=1,2,\cdots,N$), we can find $K$ nearest sketch patches from the sketches in the representation dataset within the search region around the location of $\mathbf{y}_i$. The probe sketch patch $\mathbf{y}_i$ can then be regarded as a linear combination of the $K$ nearest sketch patches $\{\mathbf{y}_{i,1},\cdots,\mathbf{y}_{i,K}\}$ weighted by a column vector $\mathbf{w}_{\mathbf{y}_i}=(w_{\mathbf{y}_{i,1}},\cdots,w_{\mathbf{y}_{i,K}})^T$. The weight vector $\mathbf{w}_{\mathbf{y}_i}$ is regarded as a representation of the probe sketch patch $\mathbf{y}_i$. For a gallery photo patch $\mathbf{x}_i^l$ from the $l$th gallery photo $\mathbf{g}^l$, where $l=1,2,\cdots,L$, we can also find $K$ nearest photo patches from the photos in the representation dataset and reconstruct the photo patch by a linear combination of these $K$ nearest photo patches weighted by $\mathbf{w}_{\mathbf{x}_i^l}$. The weight vector $\mathbf{w}_{\mathbf{x}_i^l}$ is regarded as a representation of the gallery photo patch $\mathbf{x}_i^l$. The proposed approach is based on the observation that two heterogeneous face image patches corresponding to the same location from the same person tend to have similar representations, and the representations of two heterogeneous face image patches from different persons usually differ greatly.

The reconstruction weights can be simply generated through conventional subspace learning approaches such as principal component analysis (PCA) \cite{Ref27} and locally linear embedding (LLE) \cite{Ref28}. However, these approaches neglect the spatial {structure} information which is essential for face recognition. {To this end}, we propose to utilize Markov networks to represent heterogeneous face image patches separately, which take full advantage of the spatial compatibility between adjacent patches. Once graphical representations for probe sketch patches and gallery photo patches are obtained, a CRSM to measure the similarity between the probe sketch $\mathbf{t}$ and the gallery photo $\mathbf{g}^l$ is designed. Figure \ref{Figure1} gives an overview of the proposed method. The details are introduced as follows.

\subsection{Graphical Representation}

Inspired by the successful application of Markov networks on synthesis scenarios \cite{Ref4,Ref5}, we jointly model all patches from a probe sketch or from a gallery photo on Markov networks. The joint probability of the probe sketch patches and the weights is defined as
\begin{equation}
\label{Eq:eq 1}
\begin{aligned}
&p(\mathbf{w}_{\mathbf{y}_1},\cdots,\mathbf{w}_{\mathbf{y}_N},\mathbf{y}_1,\cdots,\mathbf{y}_N)\\
=\quad&\prod_i\Phi(\mathbf{f}(\mathbf{y}_i),\mathbf{f}(\mathbf{w}_{\mathbf{y}_i}))\prod_{(i,j)\in\Xi}\Psi(\mathbf{w}_{\mathbf{y}_i},\mathbf{w}_{\mathbf{y}_j})\\
\end{aligned}
\end{equation}
where $(i,j)\in\Xi$ denotes {that} the $i$th probe sketch patch and the $j$th probe sketch patch are adjacent. {$\Xi$ represents the edge set in the sketch layer of the Markov networks.} $\mathbf{f}(\mathbf{y}_i)$ means the feature extracted from the probe sketch patch $\mathbf{y}_i$ and $\mathbf{f}(\mathbf{w}_{\mathbf{y}_i})$ denotes the linear combination of features extracted from neighboring sketch patches in the representation dataset, \textit{i.e.} $\mathbf{f}(\mathbf{w}_{\mathbf{y}_i})=\sum_{k=1}^K{w_{\mathbf{y}_{i,k}}\mathbf{f}(\mathbf{y}_{i,k})}$. $\Phi(\mathbf{f}(\mathbf{y}_i),\mathbf{f}(\mathbf{w}_{\mathbf{y}_i}))$ is the local evidence function, and $\Psi(\mathbf{w}_{\mathbf{y}_i},\mathbf{w}_{\mathbf{y}_j})$ is the neighboring compatibility function.

The local evidence function $\Phi(\mathbf{f}(\mathbf{y}_i),\mathbf{f}(\mathbf{w}_{\mathbf{y}_i}))$ is defined as
\begin{equation}
\label{Eq:eq 2}
\begin{aligned}
&\Phi(\mathbf{f}(\mathbf{y}_i),\mathbf{f}(\mathbf{w}_{\mathbf{y}_i}))\\
\propto\quad&\mbox{exp}\{-\|\mathbf{f}(\mathbf{y}_i)-\sum_{k=1}^K{w_{\mathbf{y}_{i,k}}\mathbf{f}(\mathbf{y}_{i,k})}\|^2/2\delta_\Phi^2\}\\
\end{aligned}
\end{equation}

The rationale behind the local evidence function is that $\sum_{k=1}^K{w_{\mathbf{y}_{i,k}}\mathbf{f}(\mathbf{y}_{i,k})}$ should be similar to $\mathbf{f}(\mathbf{y}_i)$. Then the weight vector $\mathbf{w}_{\mathbf{y}_i}$ is regarded as a representation of the probe sketch patch $\mathbf{y}_i$.

The neighboring compatibility function $\Psi(\mathbf{w}_{\mathbf{y}_i},\mathbf{w}_{\mathbf{y}_j})$ is defined as
\begin{equation}
\label{Eq:eq 3}
\begin{aligned}
&\Psi(\mathbf{w}_{\mathbf{y}_i},\mathbf{w}_{\mathbf{y}_j})\\
\propto\quad&\mbox{exp}\{-\|{\sum_{k=1}^K{w_{\mathbf{y}_{i,k}}\mathbf{o}_{i,k}^j}}-{\sum_{k=1}^K{w_{\mathbf{y}_{j,k}}\mathbf{o}_{j,k}^i}}\|^2/2\delta_\Psi^2\}\\
\end{aligned}
\end{equation}
where $\mathbf{o}_{i,k}^j$ represents the vector consisting of intensity values extracted from overlapping area (between the $i$th probe sketch patch and the $j$th probe sketch patch) in the $k$th nearest sketch patch of the $i$th probe sketch patch. The neighboring compatibility function is utilized to guarantee that neighboring patches have compatible overlaps.

Maximizing the joint probability function (\ref{Eq:eq 1}), we can obtain the optimal representations for the probe sketch. By substituting equations (\ref{Eq:eq 2}) and (\ref{Eq:eq 3}) into equation (\ref{Eq:eq 1}), maximizing the joint probability function (\ref{Eq:eq 1}) is equivalent to the minimization problem as follows{.}
\begin{equation}
\label{Eq:eq 4}
\begin{aligned}
\min_{\mathbf{w}}\quad&\frac{1}{2\delta_\Psi^2}\sum_{(i,j)\in\Xi}\|{\sum_{k=1}^K{w_{\mathbf{y}_{i,k}}\mathbf{o}_{i,k}^j}}-{\sum_{k=1}^K{w_{\mathbf{y}_{j,k}}\mathbf{o}_{j,k}^i}}\|^2\\
+&\frac{1}{2\delta_\Phi^2}\sum_{i=1}^N\|\mathbf{f}(\mathbf{y}_i)-\sum_{k=1}^K{w_{\mathbf{y}_{i,k}}\mathbf{f}(\mathbf{y}_{i,k})}\|^2\\
s.t.\quad&\sum_{k=1}^K{w_{\mathbf{y}_{i,k}}}=1,\ 0\leq{w_{\mathbf{y}_{i,k}}}\leq1\\
&i=1,2,\cdots,N,\ k=1,2,\cdots,K\\
\end{aligned}
\end{equation}
where $\mathbf{w}$ is the concatenation of {$\{\mathbf{w}_{\mathbf{y}_1},\cdots,\mathbf{w}_{\mathbf{y}_N}\}$} in a long-vector form. Equation (\ref{Eq:eq 4}) can be further simplified as
\begin{equation}
\label{Eq:eq 5}
\begin{aligned}
\min_\mathbf{w}\quad&\alpha\sum_{(i,j)\in\Xi}\|\mathbf{O}_i^j\mathbf{w}_{\mathbf{y}_i}-\mathbf{O}_j^i\mathbf{w}_{\mathbf{y}_j}\|^2+\sum_{i=1}^N\|\mathbf{f}(\mathbf{y}_i)-\mathbf{F}_i\mathbf{w}_{\mathbf{y}_i}\|^2
\end{aligned}
\end{equation}
where $\alpha={\delta_\Phi^2}/{\delta_\Psi^2}$. $\mathbf{F}_i$ and $\mathbf{O}_i^j$ are two matrices, with the $k$th column being $\mathbf{f}(\mathbf{y}_{i,k})$ and $\mathbf{o}_{i,k}^j$, respectively. Equation (\ref{Eq:eq 5}) can be rewritten as the following problem{.}
\begin{equation}
\label{Eq:eq 6}
\begin{aligned}
\min_\mathbf{w}\quad&\mathbf{w}^T\mathbf{Q}\mathbf{w}+\mathbf{w}^T\mathbf{c}+b\\
s.t.\quad&\sum_{k=1}^K{w_{\mathbf{y}_{i,k}}}=1,\ 0\leq{w_{\mathbf{y}_{i,k}}}\leq1,\\
&i=1,2,\cdots,N,\ k=1,2,\cdots,K\\
\end{aligned}
\end{equation}
where
\begin{equation*}
\begin{aligned}
\mathbf{Q}=&\alpha\sum_{(i,j)\in\Xi}(\mathbf{O}_i^j-\mathbf{O}_j^i)^T(\mathbf{O}_i^j-\mathbf{O}_j^i)+\sum_{i=1}^N\mathbf{F}_i^T\mathbf{F}_i\\
\mathbf{c}=&-2\sum_{i=1}^N\mathbf{F}_i^T\mathbf{f}(\mathbf{y}_i)\\
b=&\sum_{i=1}^N\mathbf{f}^T(\mathbf{y}_i)\mathbf{f}(\mathbf{y}_i)\\
\end{aligned}
\end{equation*}
The bias term $b$ has no effect on the optimization problem and we can ignore it. The problem in equation (\ref{Eq:eq 6}) is optimized by the cascade decomposition method \cite{Ref5} and then we obtain the weight matrix of the probe sketch $\mathbf{W}_{\mathbf{t}}=[\mathbf{w}_{\mathbf{y}_1},\cdots,\mathbf{w}_{\mathbf{y}_N}]$. The weight matrix $\mathbf{W}_{\mathbf{g}^l}=[\mathbf{w}_{\mathbf{x}_1^l},\cdots,\mathbf{w}_{\mathbf{x}_N^l}]$ of the $l$th gallery photo $\mathbf{g}^l$ can be obtained in a similar way as aforementioned by jointly model all the gallery photo patches from $\mathbf{g}^l$ and corresponding neighboring photo patches in the representation dataset.

To match the representation $\mathbf{w}_{\mathbf{y}_i}$ of a probe sketch patch {$\mathbf{y}_i$} to {the representation} $\mathbf{w}_{\mathbf{x}_i^l}$ {of the gallery photo patch $\mathbf{x}_i^l$, where $l=1,2,\cdots,L$}, these weight vectors are reformulated as $M$-dimensional vectors {(originally, these vectors are $K$-dimensional vectors)}. For the ease of denotations, these reformulated vectors are still represented as before. Each reformulated vector has at most $K$ nonzero values. For example, $w_{\mathbf{y}_{i,z}} (z=1,2,\cdots,M)$ is nonzero only if the $i$th patch extracted from the $z$th sketch in the representation dataset is among the $K$ nearest neighbors of the probe sketch patch $\mathbf{y}_i$.

\subsection{Coupled Representation Similarity Metric}

In order to measure the similarity between two representations $\mathbf{W}_{\mathbf{t}}$ and $\mathbf{W}_{\mathbf{g}^l}$, we calculate the similarity of each coupled patch pair respectively. Here "couple" means that the two column vectors extracted from $\mathbf{W}_{\mathbf{t}}$ and $\mathbf{W}_{\mathbf{g}^l}$ have the same column order. There are many common metric functions to measure the similarity between two vectors, such as L1 norm, L2 norm, L$\infty$ norm, the cosine distance, and the chi-square distance. However, these common metrics cannot fully exploit the characteristics of the proposed graphical representation, \emph{i.e.} two graphical representations corresponding to the same position in coupled heterogeneous face images share similar semantic meanings. For example, $w_{\mathbf{y}_{i,z}}$ and $w_{\mathbf{x}_{i,z}^l}$ represent the weights of the sketch patch and photo patch from the $z$th $(z=1,2,\cdots,M)$ sketch-photo pair in the representation dataset. Here we utilize the weights which share the same neighbors in the graphical representations to describe the semantic similarity. Inspired by the rank-based similarity measure in \cite{Ref26}, we propose a new similarity measure, namely coupled representation similarity metric (CRSM), to cater for this principle.

We compute the similarity score of the probe sketch patch $\mathbf{y}_i$ and the gallery photo patch $\mathbf{x}_i^l$ as the sum of the weights sharing the same nearest neighbors.
\begin{equation}
\label{Eq:eq 8}
\begin{aligned}
s(\mathbf{y}_i,\mathbf{x}_i^l)=0.5\sum_{z=1}^M{n_z({w}_{\mathbf{y}_{i,z}}+{w}_{\mathbf{x}_{i,z}^l})}
\end{aligned}
\end{equation}
where
\begin{equation*}
\begin{aligned}
n_z=\left\{
\begin{array}{rcl}
1, & & {w}_{\mathbf{y}_{i,z}}>0 \quad\mbox{and} \quad{w}_{\mathbf{x}_{i,z}^l}>0\\
0, & &\mbox{otherwise}\\
\end{array}
\right.
\end{aligned}
\end{equation*}

\begin{figure}[t]
\begin{center}
   \includegraphics[width=1\linewidth]{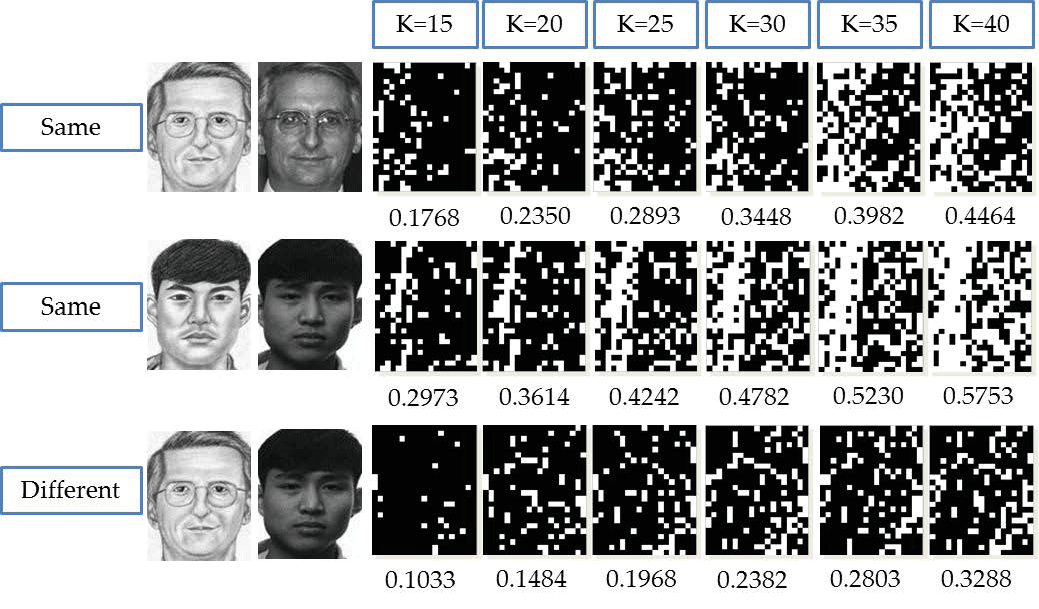}
\end{center}
   \caption{Examples of the obtained similarity map images. The left two columns show three sketch-photo pairs from the CUFSF database. The first two pairs are of the same person, and the third pair is of different persons. The corresponding similarity map images obtained are shown in the right of the sketch-photo pairs. The size of the similarity map image is the same to the face image. We have quantified the similarity map images into binary images for better visualization. The bright area indicates that the corresponding similarity score is larger than 0.5.}
\label{Figure2}
\end{figure}

The effect of the number of nearest neighbors $K$ on the similarity measurement is shown in Figure \ref{Figure2}. {The similarity map images of three sketch-photo pairs from the CUFSF database are shown as examples. The first two pairs are of the same person and the third pair is of different persons. We have quantified the similarity map images into binary images for better visualization, where the bright area denotes that the corresponding similarity score is larger than 0.5.} We find that similarity map images corresponding to heterogeneous faces of the same person tend to have more bright areas than those of different persons have. {Considering the constraints
\begin{equation*}
\begin{aligned}
&\sum_{z=1}^M{w_{\mathbf{y}_{i,z}}}=1,\quad&\sum_{z=1}^M{w_{\mathbf{x}_{i,z}^l}}=1\\
\end{aligned}
\end{equation*}
the proposed similarity measure ranges from 0 to 1.
}

The average of the similarity scores on all patch positions can be regarded as the final similarity score between the probe sketch and the gallery photo{, which is used for matching}. In Figure \ref{Figure2}, the numbers below the similarity map images are similarity scores obtained. The proposed graphical representation for HFR is summarized in \emph{Algorithm} 1 below. It should be noticed that although the process described in \emph{Algorithm} 1 does not need statistical learning, the fusion of multiple similarity metrics through statistical learning would further improve the performance, which will be shown in the experimental section.
\vspace{0.5cm}

\noindent\textbf{Algorithm 1. Graphical Representation for HFR (G-HFR)}
\begin{enumerate}
\renewcommand{\labelenumi}{\theenumi:}
\item \begin{flushleft}\textbf{Input}: representation dataset \end{flushleft} $(\mathbf{s}^1,\mathbf{p}^1),\cdots,(\mathbf{s}^M,\mathbf{p}^M)$, a probe sketch $\mathbf{t}$, gallery photos $\{\mathbf{g}^1,\cdots,\mathbf{g}^L\}$.
\item Initialize: $\mathbf{w}_{\mathbf{y}_i}=[1/K,\cdots,1/K]$, $\mathbf{w}_{\mathbf{x}_i^l}=[1/K,\cdots,1/K]$, $i=1,\cdots,N$ and $l=1,\cdots,L$; divide face images into even overlapping patches.
\item Search $K$ nearest neighbors for each probe sketch patch and gallery photo patch respectively.
\item Solve the minimization problem (\ref{Eq:eq 6}) to compute graphical representations of probe sketch $t$ and gallery photos $\{\mathbf{g}^1,\cdots,\mathbf{g}^L\}$ respectively.
\item Compute the similarity scores according to (\ref{Eq:eq 8}).
\item \textbf{Output}: the matched photo with largest similarity score.
\end{enumerate}

\section{Experiments}
\label{section IV}

In this section, we evaluated the performance of the proposed approach on four HFR scenarios tasks (viewed sketch, forensic sketch, near infrared image, and thermal infrared image). We first evaluated the effectiveness of the proposed graphical representation and the effectiveness of CRSM separately. Then we investigated the effect of different parameters and number of features on the recognition performance. Finally we validated that our approach achieved superior performance compared with state-of-the-art methods on multiple heterogeneous face databases.

\subsection{Databases}

\begin{figure}[t]
\begin{center}
   \includegraphics[width=0.9\linewidth]{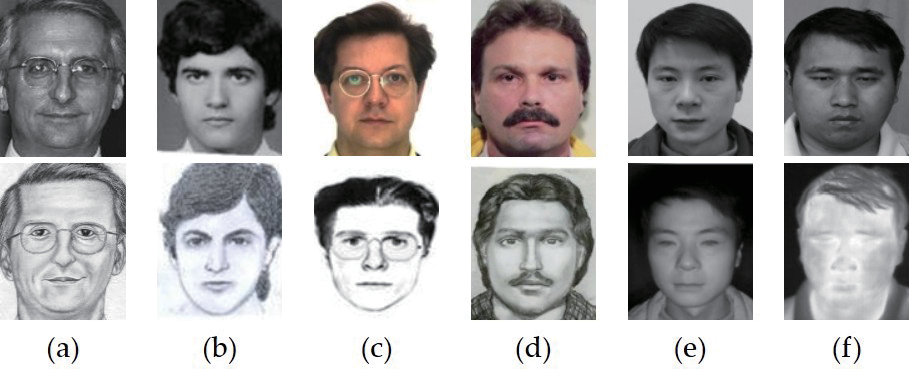}
\end{center}
   \caption{Example images of heterogeneous faces tested in this paper. (a) Viewed sketch-photo pair from the CUFSF database. (b) Semi-forensic sketch-photo pair from the IIIT-D sketch database. (c) Composite sketch-photo pair from the PRIP-VSGC database. (d) Forensic sketch-photo pair from our collected forensic sketch database. (e) Near infrared image-photo pair from the CASIA NIR-VIS 2.0 face database. (f) Thermal infrared image-photo pair from the USTC-NVIE database.}
\label{Figure3}
\end{figure}

Four different HFR scenarios are tested in this section. Example faces are shown in Figure \ref{Figure3}. Note that all the experiments are conducted with randomly partition the dataset into the representation set, the training set, and the test set. The accuracies reported in this paper are statistical results over 10 random partitions.

\subsubsection{Viewed Sketch Database}

The CUHK Face Sketch FERET Database (CUFSF) \cite{Ref18} includes 1194 sketch-photo pairs with photos collected from the FERET database \cite{Ref29}. The viewed sketches are drawn by the sketch artist when viewing the photo images. There are lighting variations in the photos and shape exaggerations in the sketches of this database. On the CUFSF database, 250 persons are randomly selected as the representation dataset, and 250 persons are randomly selected as the set for training classifiers (namely training set). The remaining 694 persons form the testing set. {Note that there is another viewed sketch database, the CUHK face sketch database (CUFS) \cite{Ref4}, which is relatively easy for state-of-the-art methods including our method to achieve accuracies higher than 99\%. Therefore, we skip over the CUFS database in this paper.}

\subsubsection{Forensic Sketch Databases}

We consider three types of forensic sketches in this paper: semi-forensic sketches, composite sketches, and forensic sketches. IIIT-D Sketch Database \cite{Ref41} contains 140 semi-forensic sketch-photo pairs with photos collected from different sources. The semi-forensic sketches are drawn based on the memory of sketch artist rather than directly viewing the photo image. The semi-forensic sketches can help bridge the gap between viewed sketches and forensic sketches. On the IIIT-D Sketch Database, the CUHK AR database \cite{Ref4} including 123 sketch-photo pairs is chosen as the representation dataset. We follow the same partition protocol in \cite{Ref45} and randomly selected 124 semi-forensic sketch-photo pairs for training the classifiers. Our collected forensic sketch database containing 168 real world forensic sketches with corresponding mug shot photos are used for test.

PRIP Viewed Software-Generated Composite Database (PRIP-VSGC) \cite{Ref42} contains 123 subjects, with photos from the AR database \cite{Ref33} and composite sketches created using FACES \cite{RefFACES} and Identi-Kit \cite{RefIdentiKit}. The composite sketches are created with facial composite software kits which synthesize a sketch by selecting a collection of facial components from candidate patterns. On the PRIP-VSGC database, we randomly selected 123 sketch-photo pairs from the CUHK Student database \cite{Ref4} to form the representation dataset. The classifiers are trained on the CUFSF database here. The 123 composite sketches generated using Identi-Kit\footnote{Currently only the 123 composite sketches generated using Identi-Kit are available in the PRIP-VSGC database.} are used for test.

Our collected forensic sketch database contains 168 real world forensic sketches and corresponding mug shot photos. The forensic sketches are drawn by sketch artists with the descriptions of eyewitnesses or victims. This database originates from a collection of images from the forensic sketch artist Lois Gibson \cite{Ref30}, the forensic sketch artist Karen Taylor \cite{Ref31}, and other internet sources. On the forensic sketch database, the CUHK AR database including 123 sketch-photo pairs is chosen as the representation dataset. We follow the same partition protocol in \cite{Ref22} and 112 persons from the forensic sketch database are randomly selected as the training set. The remaining 56 persons are used for test.

\subsubsection{Near Infrared Database}

The CASIA NIR-VIS 2.0 Face Database \cite{Ref43} contains 725 subjects, with near infrared images and photos captured by NIR and VIS cameras respectively. The age distribution of the subjects ranges from children to old people. Different from some existing methods \cite{Ref12,Ref13,Ref14} which benefit from multiple images per subject in training and gallery, only one NIR and one VIS image per subject are randomly selected in this paper to make the scenario more difficult. Therefore, there are totally 725 near infrared image-photo pairs, of which 100 pairs are randomly selected as the representation dataset. We randomly select 417 pairs to train the classifiers and the rest 208 pairs are used for test.

\subsubsection{Thermal Infrared Database}

The Natural Visible and Infrared facial Expression Database (USTC-NVIE) \cite{Ref44} contains 215 subjects, with photos captured by a visible camera and thermal infrared images captured by a infrared camera. There are illumination and facial expression variations as well as glasses disguise effect in this database. Following the same strategy with the near infrared database above, we randomly select one TIR and one VIS image per subject to make this scenario more difficult, too. There are totally 129 thermal infrared image-photo pairs\footnote{Due to the loss of some thermal and visible videos \cite{Ref44}, only 129 subjects are available in the USTC-NVIE database.}. On the thermal infrared database, 60 thermal infrared image-photo pairs are randomly selected as the representation dataset. We further randomly select 30 pairs to form the training set and the remaining 39 pairs are used for test.

\subsubsection{Enlarged Gallery}

A collection of 10,000 face photo images of 5,329 persons was used to increase the scale of the gallery, which mimic the real-world face retrieval scenarios, e.g. applications in law enforcement. The face photos in the enlarged gallery set are collected from four databases: the FERET database (2,722 photos) \cite{Ref29}, the XM2VTS database (1,180 photos) \cite{Ref34}, the CAS-PEAL database (3,098 photos) \cite{Ref49}, and the labeled faces in the wild-a (LFW-a) database (3,000 photos) \cite{Ref26}. The face images in the first three databases used are all captured under controlled conditions and their qualities are similar with those of the gallery sets in this paper. In order to increase the diversity of the enlarged gallery set, the LFW-a database is also used to construct the enlarged gallery set here. Experiments with an enlarged gallery can make results much closer to real-world HFR scenarios.

\subsection{Experimental Settings}

The parameters appeared in this paper are set as follows. A simple geometry alignment based upon five points (centers of two eyes, nose tip, left mouth corner, and right mouth corner) is performed on the face images used in this paper. These five facial points are automatically detected by the facial point detection method \cite{Ref40}, and error points are corrected manually. The only exception is that the facial points of the thermal infrared images are manually located. Each face image is cropped to $100\times125$ based on the facial points. The image patch size is {$10\times10$}, and the overlapping {area is 50\%}, \textit{i.e.} there are $P_M=456$ patches per image. The neighborhood search region is $16\times16$. In the Markov networks, we do not set $\delta_\Phi$ or $\delta_\Psi$ directly, but instead $\alpha$ is set to 0.025, where $\alpha={\delta_\Phi^2}/{\delta_\Psi^2}$. Three local descriptors, \emph{i.e.}, SURF \cite{Ref35}, SIFT \cite{Ref16}, and histograms of oriented gradients (HOG) \cite{Ref36}, are used in this paper. Each local descriptor is extracted from image patches with size of $10\times10$. For SURF, we employ the implementation embedded in the MATLAB software (available from the R2012b version), where the standard SURF-64 version was utilized. We manually set the center of the image patch as the interest point. The default parameter settings are selected and a 64-dimensional vector is returned as the SURF descriptor. For SIFT, we use an open source library \cite{Ref52}. The center of the image patch is taken as the interest point and we apply the default parameter settings to obtain the standard 128-dimensional vector. The HOG descriptor is also obtained through the open source library \cite{Ref52}. The $10\times10$ image patch is taken as the input and the cellsize is set to 5. A 124-dimensional vector is generated as the HOG descriptor. To determine other experimental settings, we conducted adjustment experiments on the CUFSF database. Once these experimental settings are determined, they are kept constant in following experiments.

For the generation of the graphical representation, the most time-consuming part lies in the neighbor searching phase and the optimization phase. For given input probe sketch patch, we first find the best match patch from each sketch in the representation dataset around the search region. Then we select $K$ most similar sketch patches as the candidates. The complexity of this process is $O(P_cP_MMP_f)$. Here $P_c$ is the number of candidates in the search region around one patch. $P_M$ is the number of patches per image. $M$ is the number of face image pairs in the representation dataset and $P_f$ is the dimensionality of the local descriptor. The optimization phase mainly depends on the number of iterations. When the iteration number is 20, it takes about 5 minutes to obtain the graphical representation of an input probe sketch from the CUFSF database. After being represented by the proposed graphical representation, the weight vector size of each image patch is $M$. Therefore, the feature dimension of graphical representation for each image is $MP_M$. The complexity of the matching process is $O(MP_M)$. In our experiments it takes about 4.2ms for one matching operation. All the experiments and computations are conducted on an Intel Core i7-4790 3.60GHz PC under MATLAB R 2012b environment.

\begin{figure}[t]
\begin{center}
\subfigure{
\label{Figure4_1}
\begin{minipage}[b]{0.47\linewidth}
    \centering
   \includegraphics[width=1.1\linewidth]{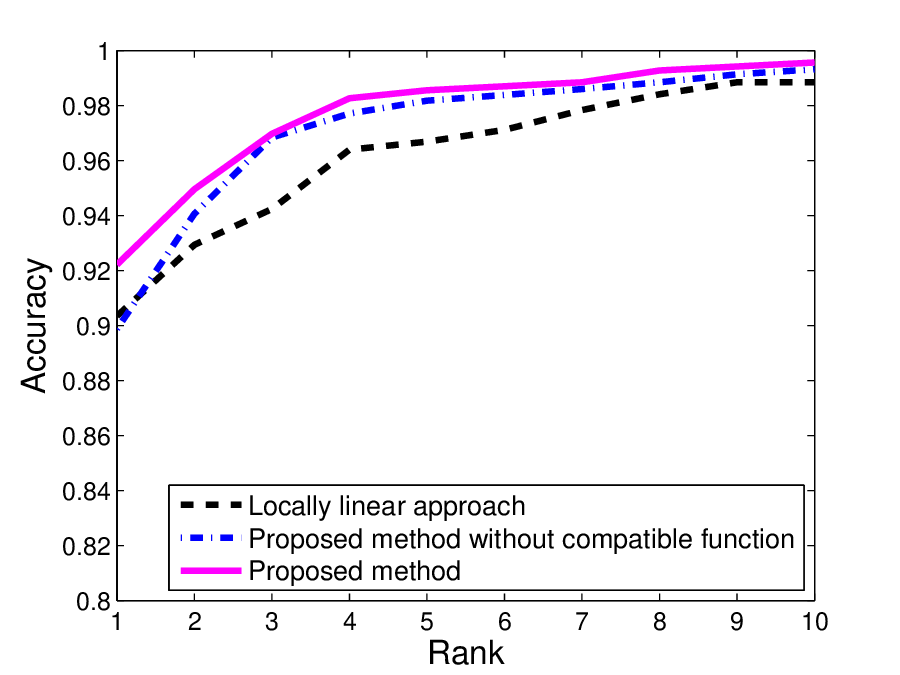}
\end{minipage}}%
\subfigure{
\label{Figure4_2}
\begin{minipage}[b]{0.47\linewidth}
    \centering
   \includegraphics[width=1.1\linewidth]{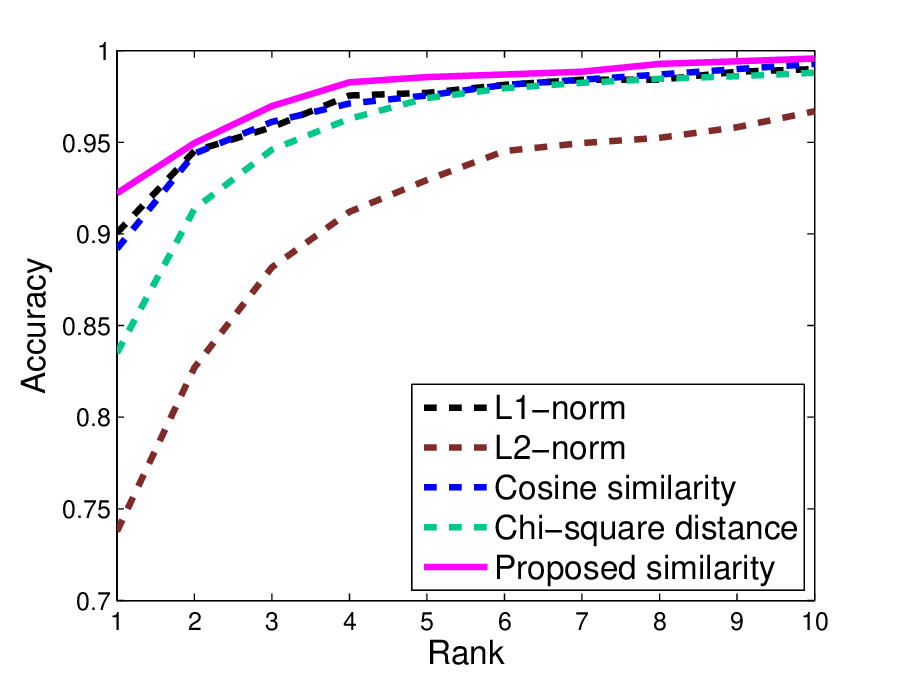}
\end{minipage}}
\subfigure{
\label{Figure4_3}
\begin{minipage}[b]{0.47\linewidth}
    \centering
   \includegraphics[width=1.1\linewidth]{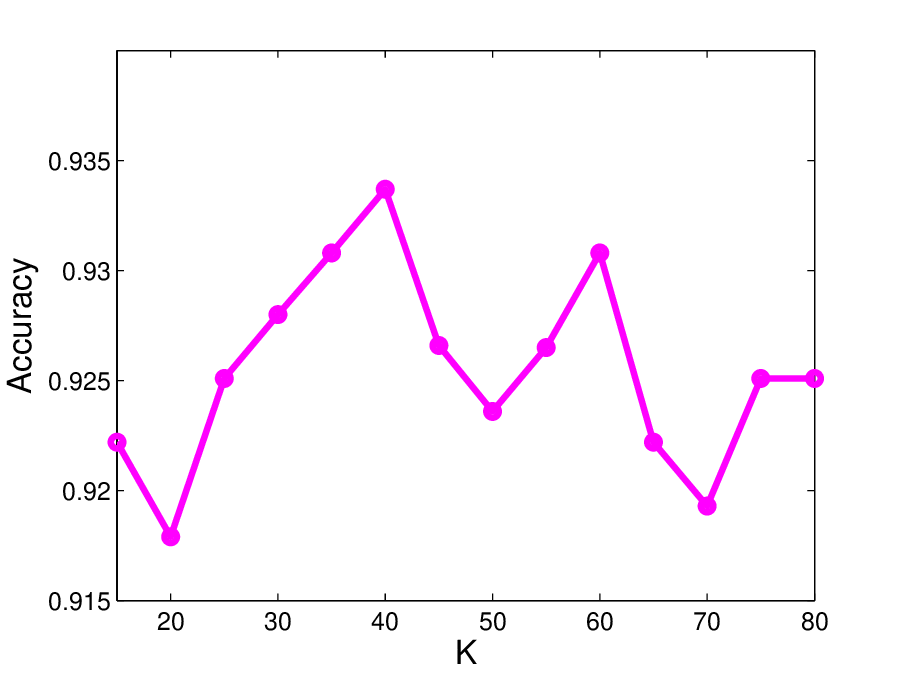}
\end{minipage}}%
\subfigure{
\label{Figure4_4}
\begin{minipage}[b]{0.47\linewidth}
    \centering
   \includegraphics[width=1.1\linewidth]{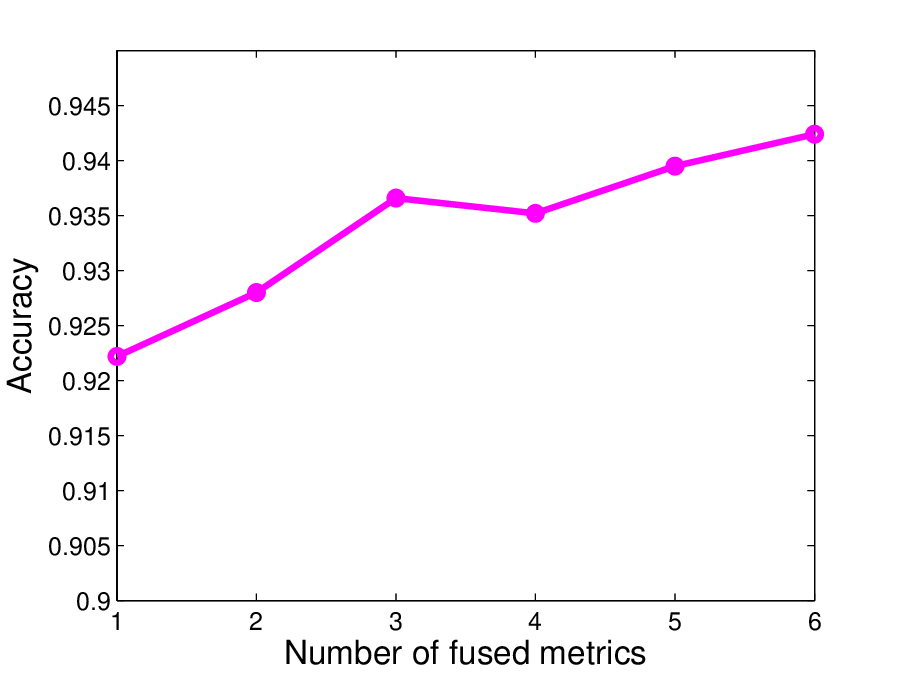}
\end{minipage}}
\end{center}
   \caption{Left top subfigure shows the evaluation for the necessity of spatial information; right top subfigure shows the comparison of the proposed CRSM with common similarity metrics; left bottom subfigure shows the accuracies of different numbers of the nearest neighbors $K$; right bottom subfigure shows the accuracies by fusion of similarity metrics. All the four experiments are conducted on the CUFSF database using the SURF feature.}
\label{Figure4}
\end{figure}

To illustrate the effectiveness of the proposed graphical representations, we first replace the Markov networks with the locally linear embedding \cite{Ref28} which ignores the spatial information. In order to better demonstrate the improvement brought by the neighboring compatibility function in equation (\ref{Eq:eq 1}), we further conduct the experiment without the compatibility function. Speeded up robust features (SURF) \cite{Ref35} are utilized as the feature descriptor and the number of the nearest neighbors $K$ is set to 15. As shown in the left top subfigure of Figure \ref{Figure4}, the spatial information is essential for HFR. By considering the relationship between neighboring patches (\emph{i.e.} taking the compatible function into consideration), the proposed method achieved superior performance.

To justify and illustrate the effectiveness of the proposed similarity metric (CRSM), we compare it with L1 norm, L2 norm L$\infty$ norm, the cosine distance, and the chi-square distance. SURF is utilized as the feature descriptor and $K$ is set to 15. The L$\infty$ norm is almost invalid on the proposed graphical representations, with a first match rate of 1.15\%. The comparison of the proposed similarity metric with other common metric functions is shown in the right top subfigure of Figure \ref{Figure4}. The L2 norm and the chi-square distance perform poorly on the proposed graphical representations. This is because these two metrics cannot exploit the characteristics of the proposed graphical representation, \emph{i.e.}, there are at most $K$ nonzero values in the $M$-dimensional vector, and simultaneously the same positions of two representation vectors in different images share similar semantic meanings. The proposed similarity measure is designed to cater for these characteristics and therefore more effective than L1 norm and cosine distance.

We evaluate the effect of the number of nearest neighbors $K$ with SURF as the feature descriptor. $K$ is set to $15,20,25,30,35,40,45,50,55,60,65,70,75$ and $80$ respectively. As shown in the left bottom subfigure of Figure \ref{Figure4}, the recognition accuracy varies with different $K$ values, and there is not a smooth relationship between $K$ and the accuracy. The rationale behind this is due to the small samples in the experiment. This inspires us to take the fusion of similarity metrics with different $K$ values which may improve the performance (actually, this point is proved in the following experiments). Considering that with the increase of $K$, more memory space is required. In the following experiments we simply set $K$ to $15,20,25,30,35$ and $40$, which is sufficient for recognition performance.

In our experiments, we find that fusion of different similarity metrics corresponding to different $K$ values would further improve the performance. We explore a linear one-class support vector machine (SVM) to fuse the similarity scores obtained by different $K$ values. We follow the fusion strategy in \cite{Ref18} and select all the intrapersonal pairs and the same number of interpersonal pairs with largest similarity scores to train the one-class SVM. As shown in the right bottom subfigure of Figure \ref{Figure4}, the increase of the number of similarity metrics does improve the recognition accuracy. The rationale behind this is that complementary information exists among different similarity metrics. Combining 6 similarity metrics increases the accuracy from 92.22\% to 94.24\%.

\begin{figure}[t]
\begin{center}
   \includegraphics[width=0.9\linewidth]{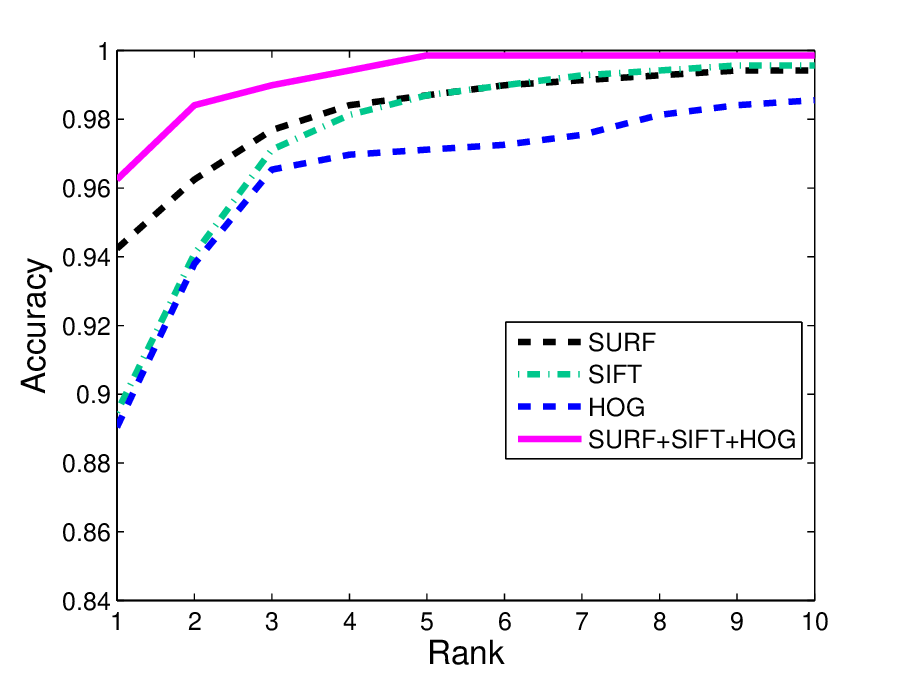}
\end{center}
   \caption{Experiments on using different features and the fusion of them on the CUFSF database.}
\label{Figure5}
\end{figure}

We also investigate the effect of the fusion of different features on the recognition performance. Because the proposed method represents the heterogeneous face images in each modality separately, common features used in homogeneous face recognition are sufficient for the task. In this paper, SURF \cite{Ref35}, SIFT \cite{Ref16}, and HOG \cite{Ref36} are employed to represent an image patch respectively. For each local descriptor, multiple graphical representations can be generated with multiple $K$ values. These graphical representations obtained based on the three descriptors are then fused through the one-class SVM, following the same strategy in \cite{Ref18}. Note that there are many other features which can also be used in the proposed method. However, since this paper mainly focuses on investigating the performance under the graphical representation framework, the selection of different types of features exceeds the scope of this work. Figure \ref{Figure5} shows that fusing models obtained from three features separately further improves the accuracy from 94.24\% (SURF), 89.48\% (SIFT), and 89.05\% (HOG) to 96.04\% respectively. This validates that fusion of the similarity metrics with different features boosts the performance.

In following experiments, G-HFR extracts three features aforementioned and 6 similarity metrics are calculated for each feature (corresponding to $K=15,20,25,30,35$ and $40$ respectively). These 18 metrics are fused by one-class SVM for final recognition task, excepted when noted.

\subsection{Experiments on the Viewed Sketch Database}

We compare the proposed G-HFR method with state-of-the-art approaches on the CUFSF database as shown in Table \ref{tab:table1}. For the transductive synthesis method (TFSPS) \cite{Ref6}, query sketches are transformed into synthesized photos, and random sampling LDA (RS-LDA) \cite{Ref37} is used to match the synthesized photos to gallery photos. Because the photos and sketches in CUFSF involve lighting variations and shape exaggerations, the synthesized photos have artifacts such as distortions. These artifacts degrade the performance of face recognition. For the common space projection based approaches PLS \cite{Ref9} and MvDA \cite{Ref11}, a discriminant common space for two modalities is learnt. Although these two approaches have a strong generality and can be applied to various heterogeneous scenarios, they perform poorly on CUFSF as shown in {Table \ref{tab:table1}}. For feature descriptor based methods LRBP \cite{Ref19} and LDoGBP \cite{Ref23}, feature descriptors which are invariant to different modalities are designed and used for recognition. These two approaches achieve good performance with accuracies of 91.12\% and 91.04\% respectively. However, these features ignore the spatial structure of faces. Our proposed method achieves a first match rate of (96.04$\pm$0.0076)\%  with 95\% confidence interval and a tenth match rate of  (99.86$\pm$0.0088)\% with 95\% confidence interval. Zhang \textit{et al.} \cite{Ref18} achieved 98.70\% verification rates (VR) at 0.1\% false acceptance rate (FAR) in comparison to 99.14\% VR at 0.1\% FAR of our proposed G-HFR method.

\begin{table}[t]
\normalsize
\begin{center}
\begin{tabular}{cc|cc}
\hline
    Method & Accuracy & Method & Accuracy \\
\hline
TFSPS \cite{Ref6} & 72.62\% & PLS \cite{Ref9} & 51\% \\
MvDA \cite{Ref11} & 55.50\% & LRBP \cite{Ref19} & 91.12\% \\
LDoGBP \cite{Ref23} & 91.04\% & G-HFR & \textbf{96.04\%} \\
\hline
\end{tabular}
\end{center}
\caption{Rank-1 recognition accuracies of the state-of-the-art approaches and our method on the CUFSF database.}
\label{tab:table1}
\end{table}

\subsection{Experiments on the Forensic Sketch Databases}

Matching forensic sketches to mug shots is much more difficult than matching aforementioned viewed sketches, because forensic sketches are drawn based on the eyewitness's descriptions. This can be easily affected by various eyewitnesses' face perceptions and sketch artists' perceptual experiences when drawing the forensic sketches. It is even harder when the eyewitness's description contains verbal overshadowing and memory distorting properties. The rank-50 accuracies of the state-of-the-art methods and the rank-50 accuracies with 95\% confidence intervals of the proposed G-HFR method on the three types of forensic sketch databases are shown in Table \ref{tab:table2}.

\begin{table*}[t]
\normalsize
\begin{center}
\begin{tabular}{ccc}
\hline
    Database & Method & Accuracy \\
\hline
IIIT-D sketch database & MCWLD \cite{Ref45} & 28.52\% \\
 & G-HFR & (30.36$\pm$0.07)\% \\
\hline
PRIP-VSGC database & Component-based \cite{Ref24} & $<$5\% \\
 & G-HFR & (51.22$\pm$0)\% \\
\hline
Forensic sketch database & P-RS \cite{Ref22} & 20.80\% \\
 & G-HFR & (31.96$\pm$0.41)\% \\
\hline
\end{tabular}
\end{center}
\caption{Rank-50 recognition accuracies of the state-of-the-art methods and rank-50 accuracies with 95\% confidence intervals of the proposed G-HFR method on three types of forensic sketch databases.}
\label{tab:table2}
\end{table*}

\begin{figure}[t]
\begin{center}
   \includegraphics[width=0.9\linewidth]{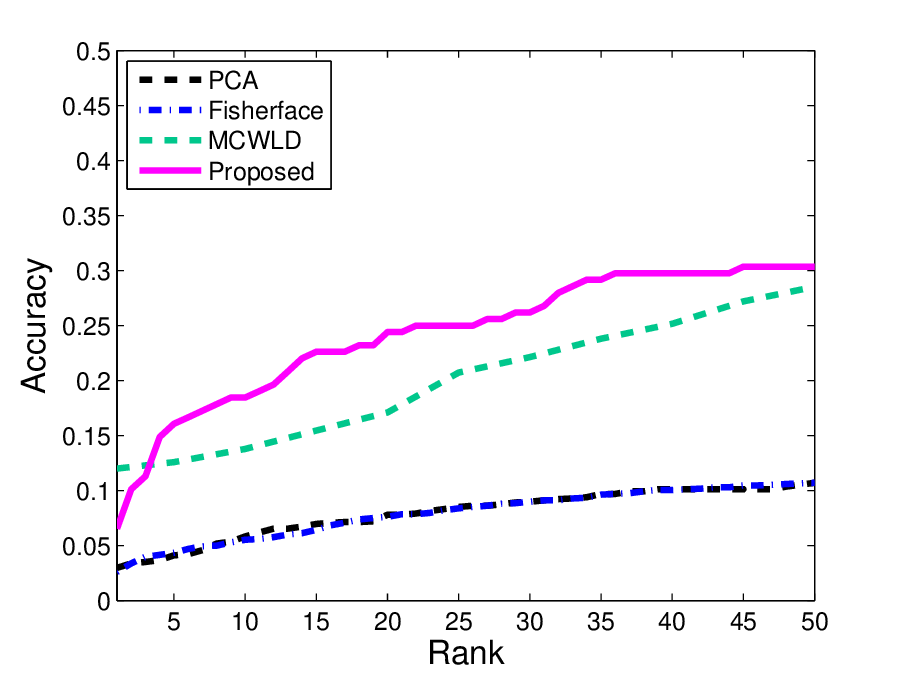}
\end{center}
   \caption{Cumulative match score comparison of the baseline methods, the MCWLD method, and our method on the IIIT-D Sketch Database.}
\label{Figure6}
\end{figure}

We first compare the recognition performance of the proposed G-HFR method with the method \cite{Ref45} on the IIIT-D Sketch Database. Considering the great differences between viewed sketch and forensic sketch, Bhatt \textit{et al.} \cite{Ref45} proposed to conduct training procedure on semi-forensic sketches and achieved better performance than the algorithm trained on viewed sketches. They encoded discriminating information from local regions using multiscale circular Weber's local descriptor (MCWLD) and optimized by an evolutionary memetic optimization algorithm. The MCWLD method utilizes 140 semi-forensic sketches for training and 190 forensic sketches are taken as the probe images. 599 face photos plus 6,324 photos form the gallery. A rank-50 accuracy of 28.52\% is achieved by this method. We follow the same partition protocol by randomly selecting 124 semi-forensic sketches for training and 168 forensic sketches are taken as the probe set. The gallery is composed of 168 mug shot photos and 10,000 photos from the enlarged gallery set. Our method achieves a rank-50 accuracy of (30.36$\pm$0.07)\% with 95\% confidence interval. To better illustrate the performance of our method, we further introduce two baseline methods (PCA \cite{Ref27} and Fisherface \cite{Ref47}) in this paper, which achieve rank-50 accuracies of (10.71$\pm$0.07)\% and (10.71$\pm$0.09)\% respectively with 95\% confidence interval on the IIIT-D Sketch Database. Figure \ref{Figure6} presents a visual comparison of cumulative match scores and shows that our method achieves superior performance under different ranks on the IIIT-D Sketch Database.

\begin{figure}[t]
\begin{center}
   \includegraphics[width=0.9\linewidth]{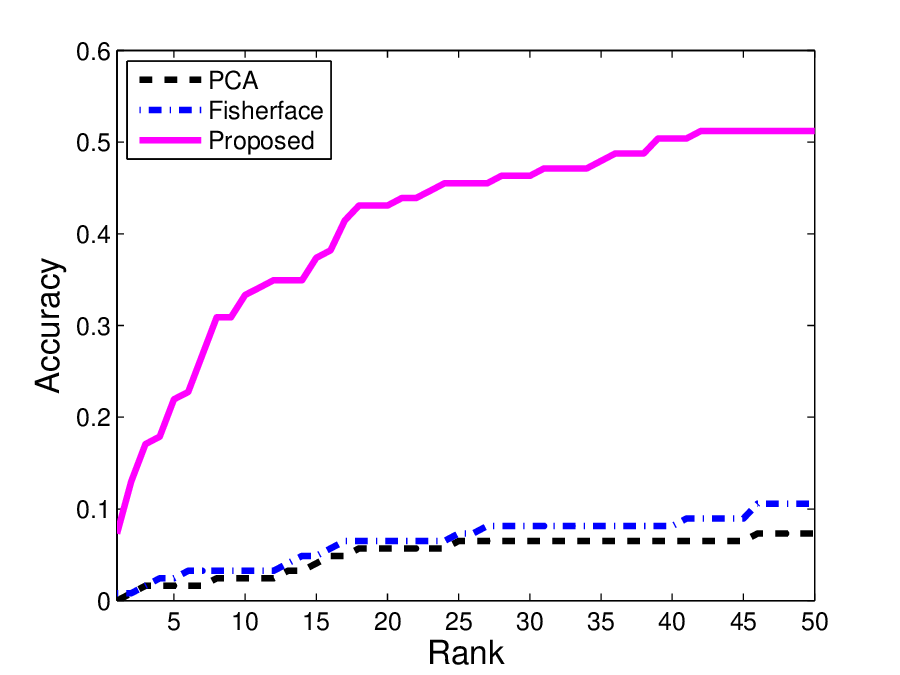}
\end{center}
   \caption{Cumulative match score comparison of the baseline methods and our method on the PRIP-VSGC Database.}
\label{Figure7}
\end{figure}

We next conduct experiment on the PRIP-VSGC database. The composite sketches are generated with each component approximated by the most similar component available in the composite software's database. Han \textit{et al.} \cite{Ref24} proposed a component-based approach by using 123 composite sketches as the probe set and 123 photos from the AR database \cite{Ref33} together with 10,000 mug shots as the gallery. Klum \textit{et al.} \cite{Ref42} recently proposed a FaceSketchID System to match facial composites with mug shots. Both the holistic and component-based algorithms in the FaceSketchID System were trained on viewed sketches and the match scores were fused to improve the performance. Note that only the 123 composite sketches generated using Identi-Kit are available in the PRIP-VSGC database, our method is evaluated on these composite sketches following the same protocol with \cite{Ref24}. The component-based approach reported their results on matching different facial components of the composite sketches generated by Identi-Kit and all the rank-50 accuracies were lower than 5\% in \cite{Ref24}. Our method achieves a rank-50 accuracy of (51.22$\pm$0)\%. Because the training and test sets are fixed on the PRIP-VSCG database, the standard deviation and 95\% confidence interval are 0 on this composite sketch database. The comparison of cumulative match scores with baseline methods is shown in Figure \ref{Figure7}.

\begin{figure}[t]
\begin{center}
   \includegraphics[width=0.9\linewidth]{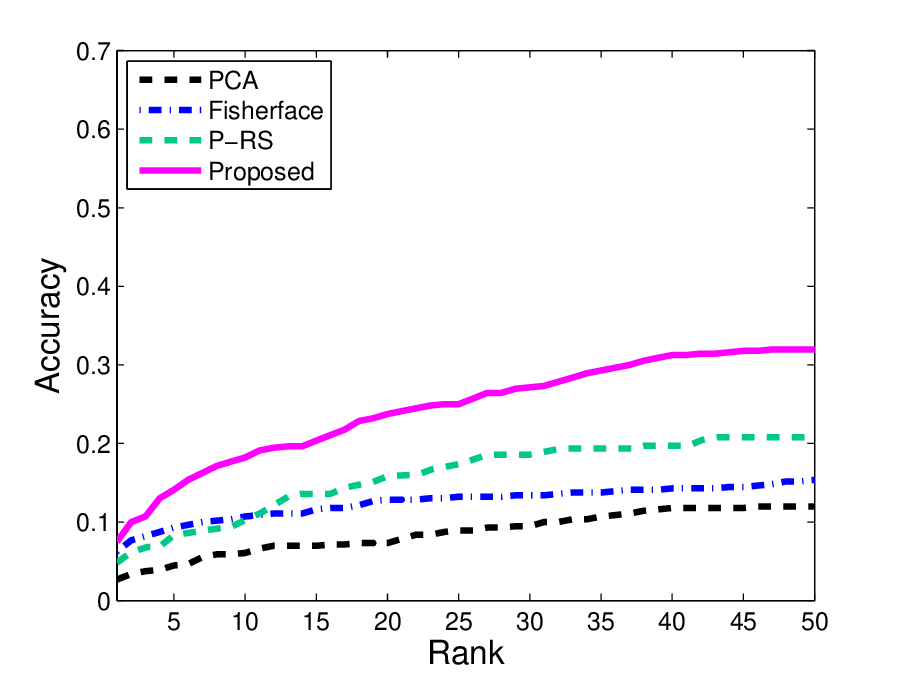}
\end{center}
   \caption{Cumulative match score comparison of the baseline methods, the P-RS method, and our method on the forensic sketch database.}
\label{Figure8}
\end{figure}

We finally conduct experiment on matching real world forensic sketches with mug shot photos. The prototype random subspaces (P-RS) method \cite{Ref22} proposed by Klare \textit{et al.} applied three different image filters and two different local feature descriptors to the probe and gallery images. A set of prototypes representing both the probe and gallery modalities are used for training and a random subspace framework is employed to boost the performance. They utilized 106 subjects for training and 53 subjects plus 10,000 mug shots for testing and achieved a rank-50 accuracy of 20.80\%. We follow the same partition protocol as in \cite{Ref22} and 112 persons randomly selected from the forensic sketch database are taken as the training set. The remaining 56 persons are used for test. The gallery set is enlarged by 10,000 photos from the enlarged gallery set. Our G-HFR method achieves a rank-50 accuracy of (31.96$\pm$0.41)\% with 95\% confidence interval, which outperforms the state-of-the-art method\cite{Ref22}. The cumulative match scores of the proposed method, the baseline methods, and the P-RS method \cite{Ref22} are shown in Figure \ref{Figure8}. Due to the small scale of available forensic sketch database, there are not enough sketches for training a strong model. It is reasonable to believe that the recognition performance can be further improved with more forensic sketches available.

\subsection{Experiments on the Near Infrared Database}

\begin{figure}[t]
\begin{center}
   \includegraphics[width=0.9\linewidth]{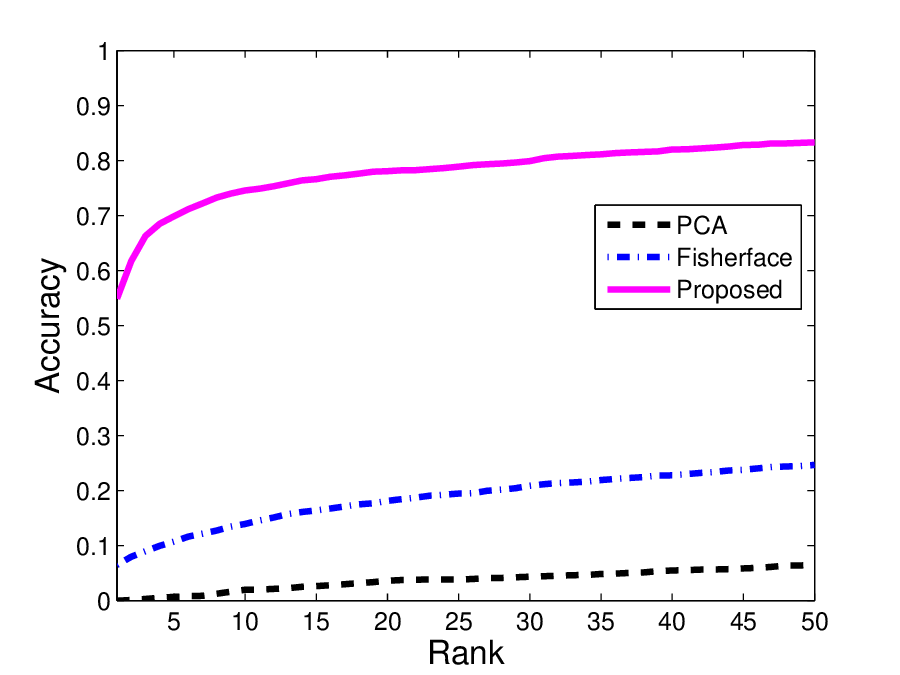}
\end{center}
   \caption{Cumulative match score comparison of the baseline methods and our method on the CASIA NIR-VIS 2.0 Face Database.}
\label{Figure9}
\end{figure}

We perform near infrared images to photos matching on the CASIA NIR-VIS 2.0 Face Database \cite{Ref43}, which is a newly constructed challenging and practical database. There are 725 subjects with 17,850 NIR and VIS images in this database. Existing NIR-VIS matching methods were trained with multiple images per subject. Motivated by \cite{Ref22}, the proposed method is trained with only one NIR-VIS pair per subject. Experiments using a smaller training set help demonstrate the value of our method. The single NIR and VIS image per subject are randomly selected. With 100 NIR-VIS pairs taken as the representation dataset, 417 NIR-VIS pairs are randomly selected as the training set and the rest 208 pairs are used for test. The gallery is enlarged by 10,000 photos from the enlarged gallery set to mimic the real-world face retrieval scenario. The proposed method achieves a rank-1 and rank-50 accuracies of (54.90$\pm$0.30)\% and (83.32$\pm$0.23)\% respectively with 95\% confidence intervals. Because this is a new database, we just compare our method with baselines. The cumulative match score comparison is shown in Figure \ref{Figure9}.

We further conduct experiments on CASIA NIR-VIS 2.0 face database by following the standard evaluation protocols provided in \cite{Ref43}. We skipped tuning the parameters on View 1 and the parameters were kept the same with the experimental settings section. We then randomly selected 150 persons from the training set on each sub-experiments of View 2 as the representation dataset. The rest NIR-VIS pairs in the training set are used for training. The testing images are still used for test following the standard evaluation protocols. The proposed method achieves a rank-1 accuracy (85.30$\pm$0.03)\% with 95\% confidence interval of. A dense SIFT with subspace LDA method proposed in \cite{Ref50} achieved a rank-1 accuracy of 73.28\%. Yi et al. \cite{Ref51} utilized restricted Boltzmann machines (RBM) to learn a shared representation for HFR and they reported an accuracy of 84.22\% by introducing the RBM and an accuracy of 86.16\% after removing the first 11 principle components of PCA.

\subsection{Experiments on the Thermal Infrared Database}

\begin{figure}[t]
\begin{center}
   \includegraphics[width=0.9\linewidth]{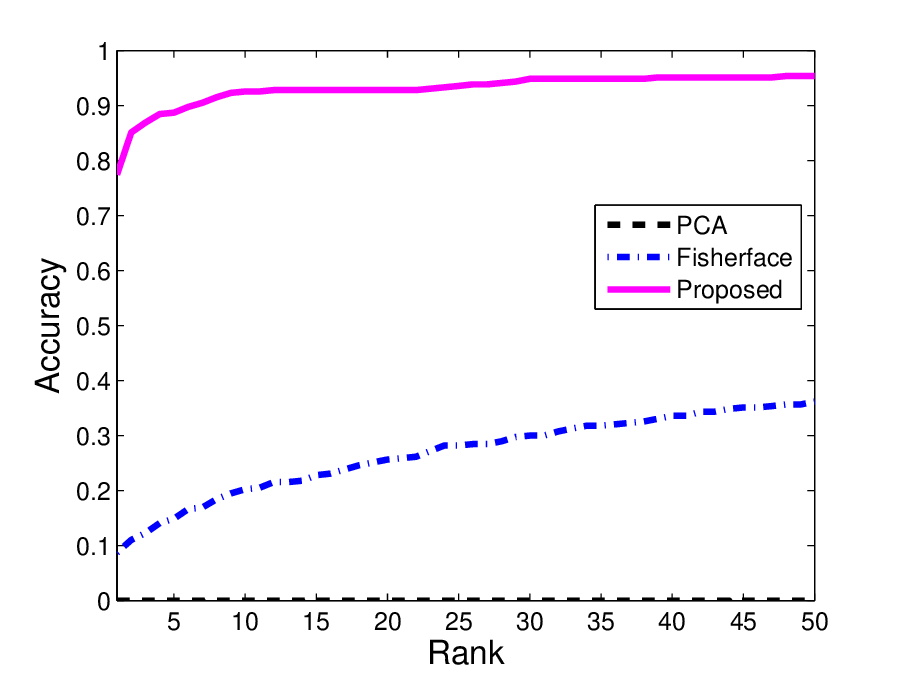}
\end{center}
   \caption{Cumulative match score comparison of the baseline methods and our method on the USTC-NVIE database.}
\label{Figure10}
\end{figure}

We perform thermal infrared images to photos matching on the USTC-NVIE database \cite{Ref44}. We randomly select one TIR image and one VIS image per subject and there are totally 129 TIR-VIS pairs. 60 TIR-VIS pairs are randomly selected as the representation dataset. We further randomly select 30 pairs to form the training set and the rest 39 pairs are used for test. The gallery is enlarged by 10,000 photos from the enlarged gallery set to make this scenario more realistic. The illumination and facial expression variations and glasses disguise effect make this database very challenging. The PCA method achieves rank-1 and rank-50 accuracies of both (0$\pm$0)\% with 95\% confidence intervals, and the Fisherface method achieves (8.72$\pm$1.09)\% and (36.15$\pm$2.50)\% respectively. Our method achieves a rank-1 and rank-50 accuracies of (77.44$\pm$2.17)\% and (95.38$\pm$0.91)\% respectively with 95\% confidence intervals. The cumulative match score comparison is shown in Figure \ref{Figure10} and our method achieves excellent performance on this scenario. To our knowledge, there are two methods performing recognition between TIR and VIS images. The {synthesis-based} TIR-VIS matching method \cite{Ref46} was evaluated on only 47 subjects in the gallery, which achieved a rank-1 accuracy of 50.06\%. The P-RS method \cite{Ref22} conducted TIR-VIS matching on a gallery of 10,333 subjects, with 667 subjects for training and 333 subjects for testing. They achieved a rank-1 accuracy of 46.7\%.

\section{Conclusions}
\label{section V}

A graphical representation based heterogeneous face recognition method (G-HFR) is proposed in this paper. G-HFR {employs} Markov networks to represent heterogeneous face images with the spatial information taken into consideration. Considering the coupled spatial property between heterogeneous face image patches, we propose a coupled representation similarity metric. Experiments are conducted to illustrate the effect of the proposed graphical representation and similarity metric in comparison to common used representations and similarity metrics. Compared with state-of-the-art methods on four heterogeneous face recognition scenarios (viewed sketch, forensic sketch, near infrared image, and thermal infrared image), G-HFR achieves superior performance in terms of face recognition accuracy. {The key benefit of the proposed G-HFR method is that the spatial information is crucial for face recognition by employing Markov networks to represent heterogeneous face images separately. The proposed graphical representation can also be applied to other fields, such as standard face recognition, facial expression recognition, and so on. In the future, the effect of more types of features would be investigated to further improve the recognition performances on each of the HFR scenarios separately. Furthermore, we would evaluate the performance of the proposed G-HFR method on more heterogeneous face recognition scenarios.}

\section*{Acknowledgments}

This work was supported in part by the National Natural Science Foundation of China under Grant 61432014, Grant 61501339, and Grant 61571343, in part by the Fundamental Research Funds for the Central Universities under Grant BDZ021403, Grant XJS15049, Grant XJS15068 and Grant JB149901, in part by Microsoft Research Asia Project based Funding under Grant FY13-RES-OPP-034, in part by the Program for Changjiang Scholars and Innovative Research Team in University of China under Grant IRT13088, in part by the Shaanxi Innovative Research Team for Key Science and Technology under Grant 2012KCT-02, and in part by the China Post-Doctoral Science Foundation under Grant 2015M580818.

\bibliographystyle{IEEEtran}
\bibliography{egbib}

\begin{IEEEbiography}[{\includegraphics[width=1in,height=1.25in]{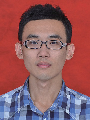}}]{Chunlei Peng}
received the B. Sc degree in electronic and information engineering from Xidian University, Xi'an, China, in 2012. He is currently pursuing the Ph.D. degree in intelligent information processing with the VIPS laboratory, School of Electronic Engineering, Xidian University, Xi'an, China. His current research interests include computer vision, pattern recognition, and machine learning.
\end{IEEEbiography}

\begin{IEEEbiography}[{\includegraphics[width=1in,height=1.25in]{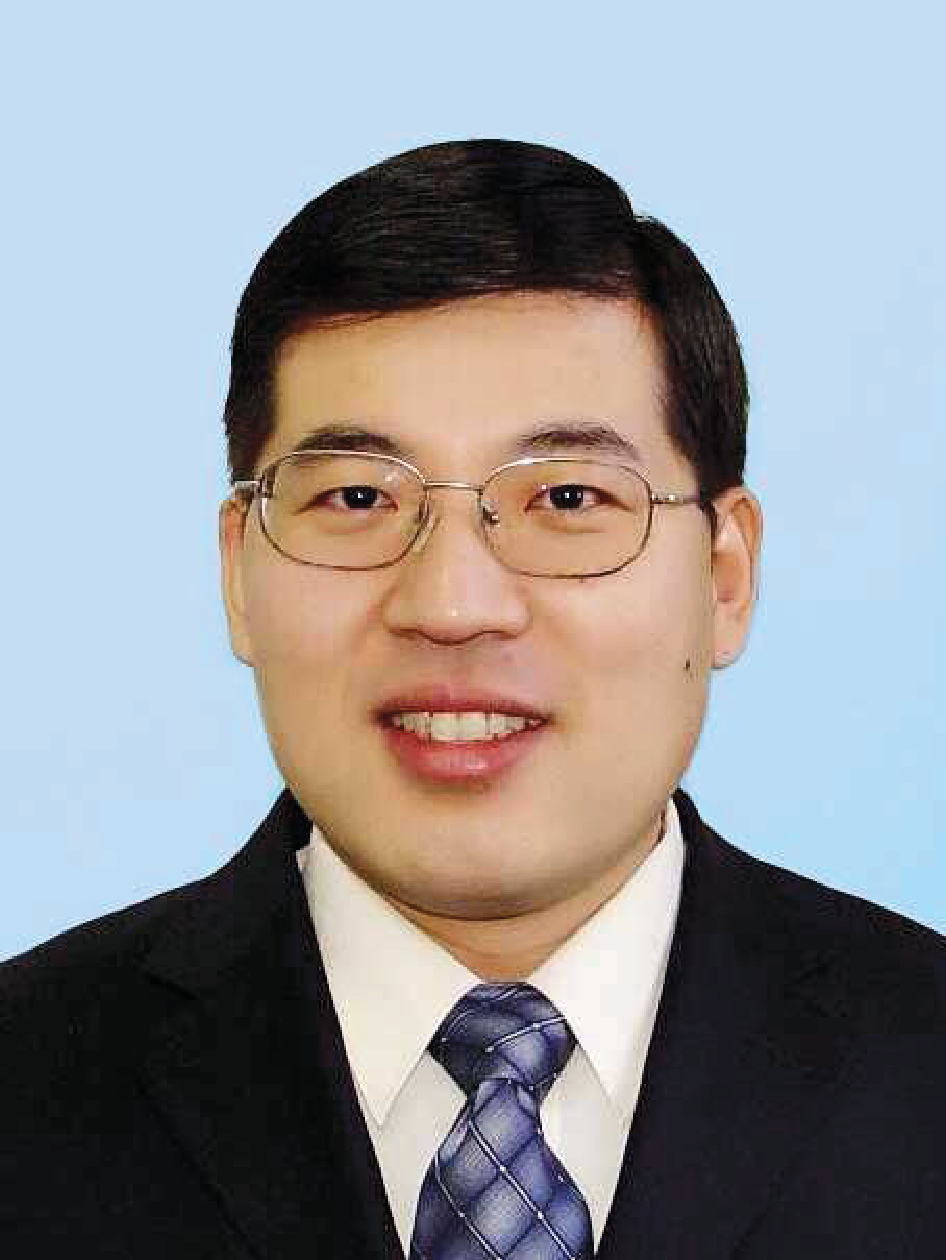}}]{Xinbo Gao} (M'02-SM'07) received the B.Eng., M.Sc., and Ph.D. degrees in signal and information processing from Xidian University, Xi'an, China, in 1994, 1997, and 1999, respectively. From 1997 to 1998, he was a Research Fellow at the Department of Computer Science, Shizuoka University, Shizuoka, Japan. From 2000 to 2001, he was a Post-doctoral Research Fellow at the Department of Information Engineering, the Chinese University of Hong Kong, Hong Kong. Since 2001, he has been at the School of Electronic Engineering, Xidian University. He is currently a Cheung Kong Professor of Ministry of Education, a Professor of Pattern Recognition and Intelligent System, and the Director of the State Key Laboratory of Integrated Services Networks, Xi'an, China. His current research interests include multimedia analysis, computer vision, pattern recognition, machine learning, and wireless communications. He has published six books and around 200 technical articles in refereed journals and proceedings. Prof. Gao is on the Editorial Boards of several journals, including Signal Processing (Elsevier), and Neurocomputing (Elsevier). He served as the General Chair/Co-Chair, Program Committee Chair/Co-Chair, or PC Member for around 30 major international conferences. He is a fellow of the Institution of Engineering and Technology and a fellow of Chinese Institute of Electronics.
\end{IEEEbiography}

\begin{IEEEbiography}[{\includegraphics[width=1in,height=1.25in]{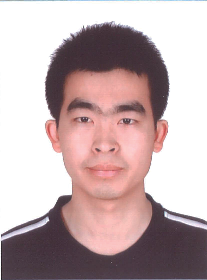}}]{Nannan Wang} (M'15) received the B. Sc degree in information and computation science from Xi¡¯an University of Posts and Telecommunications in 2009. He received his Ph.D. degree in information and telecommunications engineering in 2015. Now, he works with the state key laboratory of integrated services networks at Xidian University. From September 2011 to September 2013, he has been a visiting Ph.D. student with the University of Technology, Sydney, NSW, Australia. His current research interests include computer vision, pattern recognition, and machine learning. He has published more than 10 papers in refereed journals and proceedings including International Journal of Computer Vision (IJCV), IEEE T-NNLS, T-IP, T-CSVT etc.
\end{IEEEbiography}

\begin{IEEEbiography}[{\includegraphics[width=1in,height=1.25in]{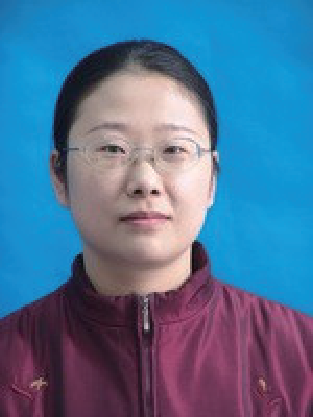}}]{Jie Li} received the B.Sc. degree in electronic engineering, the M.Sc. degree in signal and information processing, and the Ph.D. degree in circuit and systems, from Xidian University, Xi'an, China, in 1995, 1998, and 2004, respectively. She is currently a Professor in the School of Electronic Engineering, Xidian University, China. Her research interests include image processing and machine learning. In these areas, she has published around 50 technical articles in refereed journals and proceedings including IEEE T-IP, T-CSVT, Information Sciences etc.
\end{IEEEbiography}

\end{document}